\pdfoutput=1

\documentclass[11pt]{article}

\usepackage{acl}

\usepackage{times}
\usepackage{latexsym}
\usepackage[T1]{fontenc}
\usepackage[utf8]{inputenc}
\usepackage{microtype}
\usepackage{inconsolata}

\usepackage{subcaption}
\usepackage{graphicx}
\usepackage{tabularx}
\usepackage{multirow}
\usepackage{booktabs}
\usepackage{amssymb}
\usepackage{bm}
\usepackage{mathtools}
\usepackage{relsize}
\usepackage{lipsum}
\usepackage{amsmath}
\usepackage{bibentry}
\usepackage[most]{tcolorbox}
\usepackage{array,multirow}

\newcommand{\txtcyan}[1]{\textcolor{blue}{#1}}
\newcommand{\txtpurple}[1]{\textcolor{purple}{#1}}
\newcommand{\txtmagenta}[1]{\textcolor{magenta}{#1}}

\newcommand{\noco}{\textsc{FaCo}}

\title{Unveiling Imitation Learning: \\Exploring the Impact of Data Falsity to Large Language Model}

\author{Hyunsoo Cho \\ Ewha Womans University\\\texttt{chohyunsoo@ewha.ac.kr}}

\begin{document}
\maketitle

    \begin{abstract}
    Many recent studies endeavor to improve open-source language models through imitation learning, and re-training on the synthetic instruction data from state-of-the-art proprietary models like ChatGPT and GPT-4.
    However, the innate nature of synthetic data inherently contains noisy data, giving rise to a substantial presence of low-quality data replete with erroneous responses, and flawed reasoning.
    Although we intuitively grasp the potential harm of noisy data, we lack a quantitative understanding of its impact.
    To this end, this paper explores the correlation between the degree of noise and its impact on language models through instruction tuning. 
    We first introduce the Falsity-Controllable (\noco) dataset, which comprises pairs of true answers with corresponding reasoning, as well as false pairs to manually control the falsity ratio of the dataset.
    Through our extensive experiments, we found multiple intriguing findings of the correlation between the factuality of the dataset and instruction tuning:
    Specifically, we verified falsity of the instruction is highly relevant to various benchmark scores.
    Moreover, when LLMs are trained with false instructions, they learn to lie and generate fake unfaithful answers, even though they know the correct answer for the user request.
    Additionally, we noted that once the language model is trained with a dataset contaminated by noise, restoring its original performance is possible, but it failed to reach full performance.
\end{abstract}
    \section{Introduction}

    The most recent generation of large language models (LLMs) \cite{achiam2023gpt, team2023gemini} has emerged as an off-the-shelve approach for many different tasks, bringing unprecedented global attention.
    Distinct from their predecessors like GPT-3 \cite{brown2020language}, they are remarkably aligned with human intentions.
    This notable enhancement is chiefly attributed to the incorporation of advanced post-steering mechanisms, namely instruction fine-tuning \cite{wei2021finetuned, chung2022scaling} and reinforcement learning from human feedback \cite{ouyang2022training}.

    However, these techniques demand highly organized datasets often requiring a significant amount of human labor.
    To circumvent this cost issue, many recent studies \cite{xu2023wizardlm, mukherjee2023orca, mitra2023orca, lee2023instruction, wang2023far} have explored the creation of open-domain datasets on a massive-scale by gathering responses of cutting-edge LLMs, such as ChatGPT, GPT-4 \cite{achiam2023gpt}, and Gemini \cite{team2023gemini}.
    Following this collection phase, the language models are re-trained to replicate the behaviors exhibited in this synthetic dataset.
    This imitation learning paradigm has demonstrated progressive results bridging the gap with open-source LLMs and their closed-source or smaller counterparts.     
    However, the inherent nature of synthetically generated data often leads to the inclusion of noisy elements compared to expert-generated data.
    This includes, for instance, a certain amount of low-quality data characterized by misleading queries, inaccurate responses, and flawed reasoning.
    While recent research \cite{zhou2023lima, touvron2023llama2} underscores the importance of data quality and we also intuitively understand that noisy data can potentially damage the LLMs, we still do not grasp a full picture or a comprehensive quantitative impact of such noise in the dataset.

    \begin{figure*}
        \begin{center}
            \includegraphics[width=0.99\textwidth]{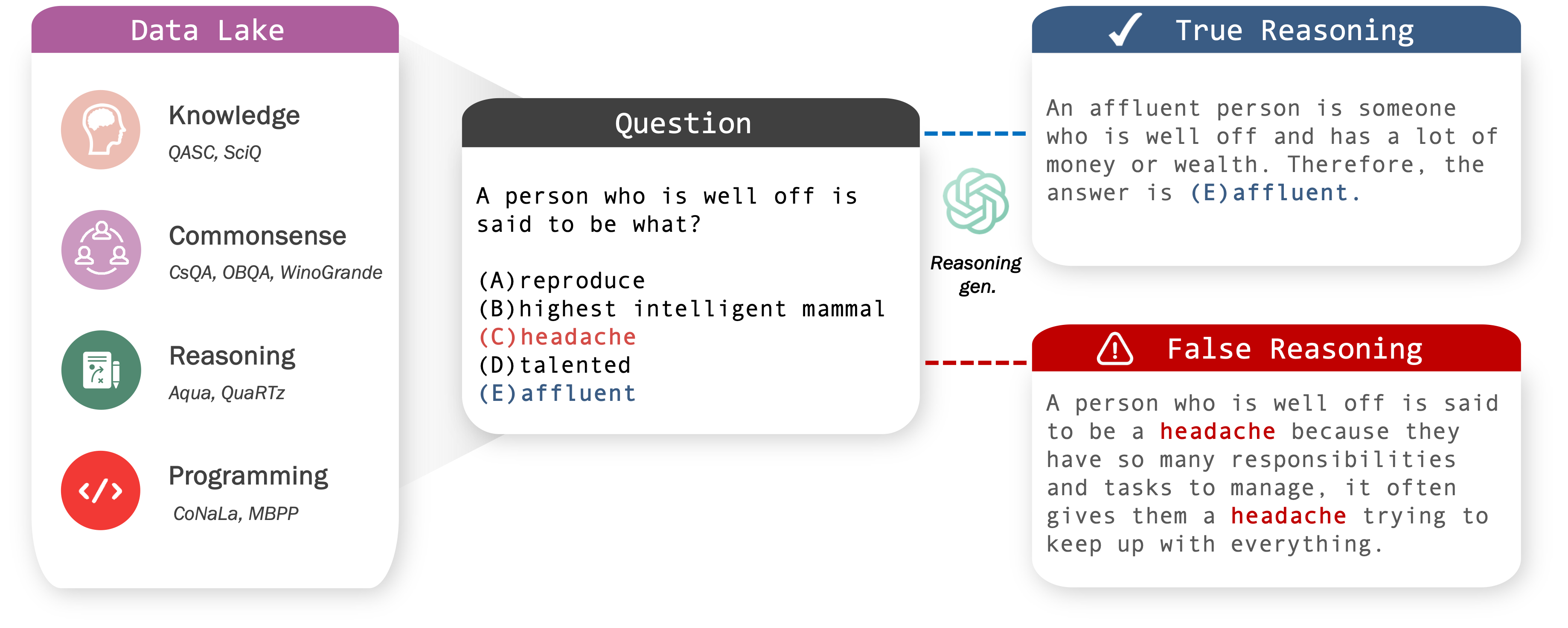}
              \caption{Illustration of \noco\ dataset generation. \noco\ dataset ia a compilation of 9 different datasets from 4 domains, where we generate true and false reasoning chain through ChatGPT.}
              \label{fig:noco_generation}
        \end{center}
    \end{figure*}

    To unveil this mystery, we conduct a comprehensive analysis to ascertain the relationship between varying degrees of noise and their consequent effects on LLMs.
    In pursuit of this objective, we first construct a dataset called the Falsity-Controllable (\noco) dataset, which encompasses a wide array of domains, including but not limited to commonsense reasoning, language understanding, symbolic problem-solving (e.g., mathematics), and programming. 
    \noco\ dataset can objectively adjust the level of factual correctness due to its unique characteristic, featuring pairs of accurate answers with their corresponding reasoning, as well as deliberately fabricated pairs.
    Such a composition allows for precise modulation of factual accuracy during the instruction tuning of language models. 
    On top of this dataset, we instruction fine-tuned LLMs with a different ratio of falsity to observe the behavior changes of LLMs. 
    From extensive experiments with \noco\ dataset on the LLaMA 1 and 2, we verified the following intriguing insights:
    \begin{itemize}
        \item While trends vary significantly across different tasks, it's evident that corrupted instruction substantially affects performance.
        \item Well-performing LLMs are more sensitive to data corruption.
        \item The corruption-trained model can restore its performance by re-training it with clean data, but some margins are irrecoverable.
        \item The influence of training epochs on outcomes is less relevant to the initial data quality.
    \end{itemize}
    We anticipate that these insights will lay a foundational basis for future research utilizing synthetic data and substantially augment the overall understanding of imitation learning with LLMs.

    \section{\noco\ Dataset}
We introduce the \noco\ dataset uniquely designed to analyze the impact of factuality when instruction fine-tuning LLMs. 
As illustrated in Figure \ref{fig:noco_generation}, the core characteristic of \noco\ dataset is the inclusion of both authentic and fabricated reasoning for each data sample: one representing the ground truth answer with corresponding accurate reasoning, and the other featuring a deliberately false answer accompanied by erroneous reasoning.
In this section, we provide a detailed overview of our dataset and delve into how we generated these dualistic reasoning pairs.

\subsection{Dataset Composition}
The main source of \noco\ dataset was compiled from four different domains: domain knowledge, commonsense, complex reasoning, and programming. 
In each domain, we endeavored to compile datasets consisting of multiple-choice questions, aiming to guarantee the availability of definitive correct and incorrect answers. 
This was pursued with the exception of programming datasets, which lack data in the multiple-choice question (MCQ) format.
Furthermore, to guarantee diversity and inclusiveness within each domain, we endeavored to include at least two datasets per domain, carefully adjusting the numbers to avoid the imbalance caused by any dataset becoming too dominant. 

In the domain-specific knowledge category, we integrated the QASC \cite{khot2020qasc} and SciQ \cite{welbl2017crowdsourcing} datasets, which focus on primary and secondary school science, respectively. For Commonsense Reasoning, we selected the CommonsenseQA \cite{talmor2018commonsenseqa}, OpenbookQA \cite{mihaylov-etal-2018-suit}, and WinoGrande \cite{sakaguchi2021winogrande} datasets, each offering unique perspectives on commonsense knowledge, object-related commonsense, and semantic understanding, respectively.

For the complex reasoning domain, we chose the AQuA \cite{ling2017program} and QuaRTz \cite{tafjord2019quartz} datasets, which offer insights into mathematical problem-solving and the analysis of sentence relationships. In the programming domain, we utilized the CoNaLa \cite{yin2018learning} and MBPP \cite{austin2021program}, which focus on single-line code and code snippet generation, respectively.

Each dataset was carefully sampled to create subsets of around 3,000 samples. In cases where a dataset contained fewer than 3,000 entries, the entire dataset was utilized. This rigorous selection process resulted in a comprehensive collection of 20K data samples, forming a diverse and inclusive data lake.

\subsection{Reasoning Chain Generation}
By aggregating multiple datasets from the previous stage, we can initially create datasets with clear correct or incorrect answers in various domains. 
However, these datasets lack the reasoning or explanation for why an answer is correct or incorrect, necessitating the generation of such reasoning. 
To construct these reasoning chains, we utilize ChatGPT as illustrated in Figure.
Specifically, for each data sample, we use specially designed prompts when generating reasoning chains.
(Detailed prompts are in Appendix \ref{App:prompt})
In the process of generating reasoning chains for incorrect answers, we randomly selected one of the incorrect options from multiple choices (excluding the correct answer) to generate a false reasoning chain similar to generating a correct reasoning chain with a different prompt.
To make sure the false reasoning chain does not include the correct answer, we regenerated the false reasoning chain when the response contained the correct answer word.
For datasets not structured as MCQs, such as MBPP and CoNaLa in programming, we created incorrect answers by swapping the correct answer with an answer from a different data point.
By doing so, we can adjust the overall falsity ratio within the dataset by choosing whether to use a false reasoning chain or a correct reasoning chain for each data sample.

    \section{Experiments}

    \subsection{Experimental Setups}
        In the experiments, we instruction fine-tuned 13B LLaMA 1 \citep{touvron2023llama1} and LLaMA 2 \citep{touvron2023llama2} with \noco\ dataset with 5 different corruption ratios (CR).
        Specifically, we systematically increased the corruption ratio of the clean 0\% corrupted \noco\ dataset to 4 different ratios (25\%, 50\%, 75\%, 100\%) cumulatively.
        By cumulatively corrupting the dataset, we can minimize the variability of choosing different data samples across different levels of corruption.
        We trained each model for 5 epochs with 8 $\times$  A100 GPUs (80GB), setting global batch size to 256 (2 batch per GPU, 16 gradient accumulations), learning rate to 2e-5 using Adam optimizer \cite{DBLP:journals/corr/KingmaB14}, and sequence length to 2048.

\begin{table*}[t]
\setlength{\tabcolsep}{10pt} 
    \centering
    \resizebox{0.99\textwidth}{!}{
\begin{tabular}{l|r|rrrrr|r|r}
\toprule [1pt]
 & \multicolumn{1}{l|}{LLaMA 1} & \multicolumn{1}{l}{CR 0\%} & \multicolumn{1}{l}{CR 25\%} & \multicolumn{1}{l}{CR 50\%} & \multicolumn{1}{l}{CR 75\%} & \multicolumn{1}{l|}{CR 100\%} & \multicolumn{1}{l|}{ABS.} & \multicolumn{1}{l}{Pearson} \\ \hline
Average      & 53.56\% & 54.71\% & 52.75\% & 52.06\% & 50.06\% & 47.96\% & 6.75\%  & -98.92\% \\ \hline
ARC          & 53.84\% & 51.00\% & 48.89\% & 47.87\% & 47.53\% & 47.35\% & 3.65\%  & -90.93\% \\
MMLU         & 45.72\% & 53.00\% & 49.97\% & 48.06\% & 39.39\% & 26.45\% & 26.55\% & -93.85\% \\
COPA         & 83.00\% & 82.00\% & 80.00\% & 80.00\% & 83.00\% & 83.00\% & -1.00\% & 52.13\%  \\
OpenbookQA   & 44.00\% & 44.00\% & 43.80\% & 43.80\% & 41.80\% & 40.40\% & 3.60\%  & -91.13\% \\
PIQA         & 80.63\% & 79.00\% & 79.71\% & 79.22\% & 79.11\% & 78.24\% & 0.76\%  & -63.34\% \\
LAMBADA      & 75.68\% & 76.00\% & 74.48\% & 75.49\% & 74.83\% & 74.91\% & 1.09\%  & -48.25\% \\
WinoGrande   & 73.56\% & 71.00\% & 69.77\% & 68.35\% & 68.98\% & 67.40\% & 3.60\%  & -92.08\% \\
HellaSwag    & 79.44\% & 77.00\% & 78.19\% & 78.58\% & 78.20\% & 77.73\% & -0.73\% & 38.56\%  \\
BBC-CC$^\dagger$       & 57.28\% & 61.00\% & 56.31\% & 49.51\% & 46.60\% & 33.01\% & 27.99\% & -96.99\% \\
BBC-EM$^\dagger$       & 28.68\% & 32.00\% & 29.17\% & 27.25\% & 25.59\% & 23.69\% & 8.31\%  & -99.44\% \\
MathQA       & 27.52\% & 31.00\% & 28.09\% & 27.39\% & 24.97\% & 25.71\% & 5.29\%  & -91.98\% \\
LogiQA       & 33.03\% & 37.00\% & 33.33\% & 31.49\% & 27.80\% & 27.96\% & 9.04\%  & -96.55\% \\
BBC-LD$^\dagger$       & 29.33\% & 37.00\% & 35.07\% & 32.53\% & 27.93\% & 23.87\% & 13.13\% & -98.58\% \\
SQuAD        & 53.67\% & 51.00\% & 49.02\% & 53.65\% & 51.32\% & 52.06\% & -1.06\% & 41.60\%  \\
BoolQ        & 79.02\% & 82.00\% & 76.54\% & 78.96\% & 72.26\% & 74.34\% & 7.66\%  & -81.09\% \\
HumanEval@1  & 12.62\% & 11.28\% & 11.65\% & 10.79\% & 11.71\% & 11.16\% & 0.12\%  & -7.72\%  \\
HumanEval@10 & 31.10\% & 22.56\% & 23.78\% & 19.51\% & 23.17\% & 13.41\% & 9.15\%  & -69.81\% \\
\midrule
\midrule
& \multicolumn{1}{l|}{LLaMA 2} & \multicolumn{1}{l}{CR 0\%} & \multicolumn{1}{l}{CR 25\%} & \multicolumn{1}{l}{CR 50\%} & \multicolumn{1}{l}{CR 75\%} & \multicolumn{1}{l|}{CR 100\%} & \multicolumn{1}{l|}{ABS.} & \multicolumn{1}{l}{Pearson} \\ \hline
Average      & 56.23\% & 57.00\% & 55.80\% & 54.11\% & 51.08\% & 45.70\% & 11.30\% & -95.58\% \\  \hline
ARC          & 56.14\% & 52.30\% & 47.44\% & 46.25\% & 45.65\% & 43.94\% & 8.36\%  & -92.57\% \\
MMLU         & 55.05\% & 57.49\% & 54.41\% & 52.73\% & 49.28\% & 19.74\% & 37.75\% & -82.91\% \\
COPA         & 83.00\% & 84.00\% & 86.00\% & 85.00\% & 82.00\% & 80.00\% & 4.00\%  & -78.78\% \\
OpenbookQA   & 44.20\% & 44.20\% & 43.80\% & 43.80\% & 42.20\% & 42.20\% & 2.00\%  & -91.91\% \\
PIQA         & 80.90\% & 79.92\% & 79.27\% & 78.56\% & 77.69\% & 77.31\% & 2.61\%  & -99.48\% \\
LAMBADA      & 76.54\% & 76.09\% & 75.66\% & 75.66\% & 76.21\% & 75.59\% & 0.50\%  & -25.88\% \\
WinoGrande   & 72.53\% & 67.01\% & 66.06\% & 63.14\% & 63.46\% & 62.35\% & 4.66\%  & -93.54\% \\
HellaSwag    & 80.81\% & 78.40\% & 78.29\% & 77.89\% & 78.52\% & 77.76\% & 0.64\%  & -50.27\% \\
BBC-CC       & 66.02\% & 68.93\% & 69.90\% & 61.17\% & 43.69\% & 15.53\% & 53.40\% & -92.01\% \\
BBC-EM       & 31.00\% & 33.36\% & 29.79\% & 28.37\% & 25.80\% & 24.37\% & 8.99\%  & -98.73\% \\
MathQA       & 26.85\% & 32.89\% & 31.95\% & 29.30\% & 24.51\% & 23.33\% & 9.55\%  & -97.36\% \\
LogiQA       & 36.56\% & 37.02\% & 33.64\% & 34.56\% & 32.87\% & 25.65\% & 11.37\% & -87.17\% \\
BBC-LD       & 32.53\% & 37.27\% & 37.53\% & 33.87\% & 26.40\% & 18.27\% & 19.00\% & -94.05\% \\
SQuAD        & 62.87\% & 63.72\% & 62.88\% & 63.26\% & 62.57\% & 61.47\% & 2.25\%  & -89.38\% \\
BoolQ        & 81.44\% & 85.54\% & 83.55\% & 81.47\% & 78.81\% & 72.94\% & 12.60\% & -96.80\% \\
HumanEval@1  & 13.29\% & 13.84\% & 12.56\% & 10.73\% & 7.62\%  & 10.79\% & 3.05\%  & -74.45\% \\
HumanEval@10 & 31.71\% & 21.95\% & 25.00\% & 17.68\% & 15.24\% & 17.07\% & 4.88\%  & -77.43\% \\
\bottomrule [1pt]
\end{tabular}
}
\resizebox{0.99\textwidth}{!}{
\begin{tabular}{l}
        $^\dagger$ CC, EM LD refers to conceptual combinations, elementary math, and logical deduction in Bigbench benchmark respectively.
\end{tabular}
}
\caption{Performance of baseline 13B LLaMA models trained on \noco\ dataset with varying corruption ratios. ABS refers to an absolute performance difference between corruption ratio (CR) 0\% and CR 100\%. Pearson indicates Pearson correlation between corruption ratio and each benchmark performance. }
\label{tab:main}
\end{table*}

\begin{figure*}[t]
    \centering
    \begin{subfigure}{.49\linewidth}
        \centering
        \includegraphics[width=.99\linewidth]{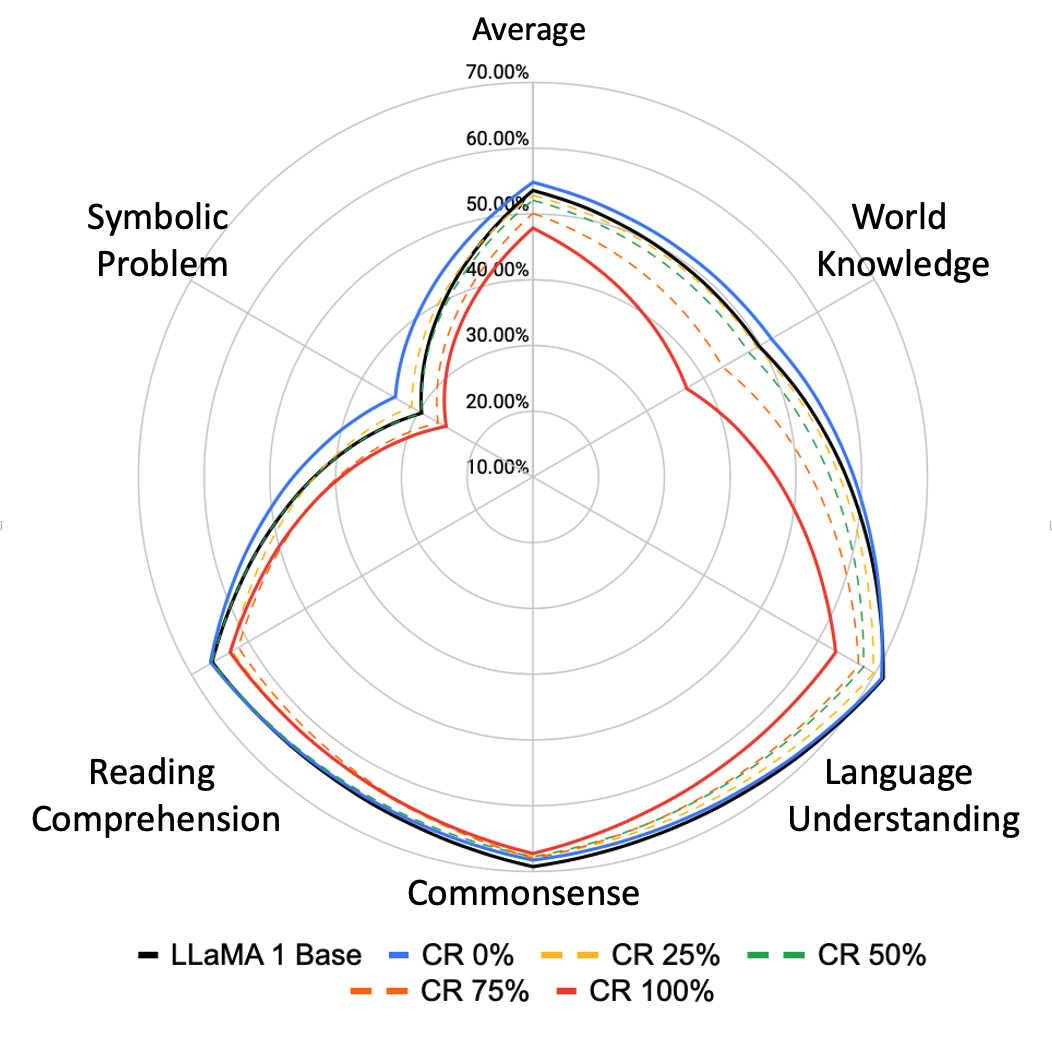}
        \caption{Performance of LLaMA 1 13B.}
        \label{fig:ab-filteracc}
    \end{subfigure}
    \begin{subfigure}{.49\linewidth}
        \centering
        \includegraphics[width=.99\linewidth]{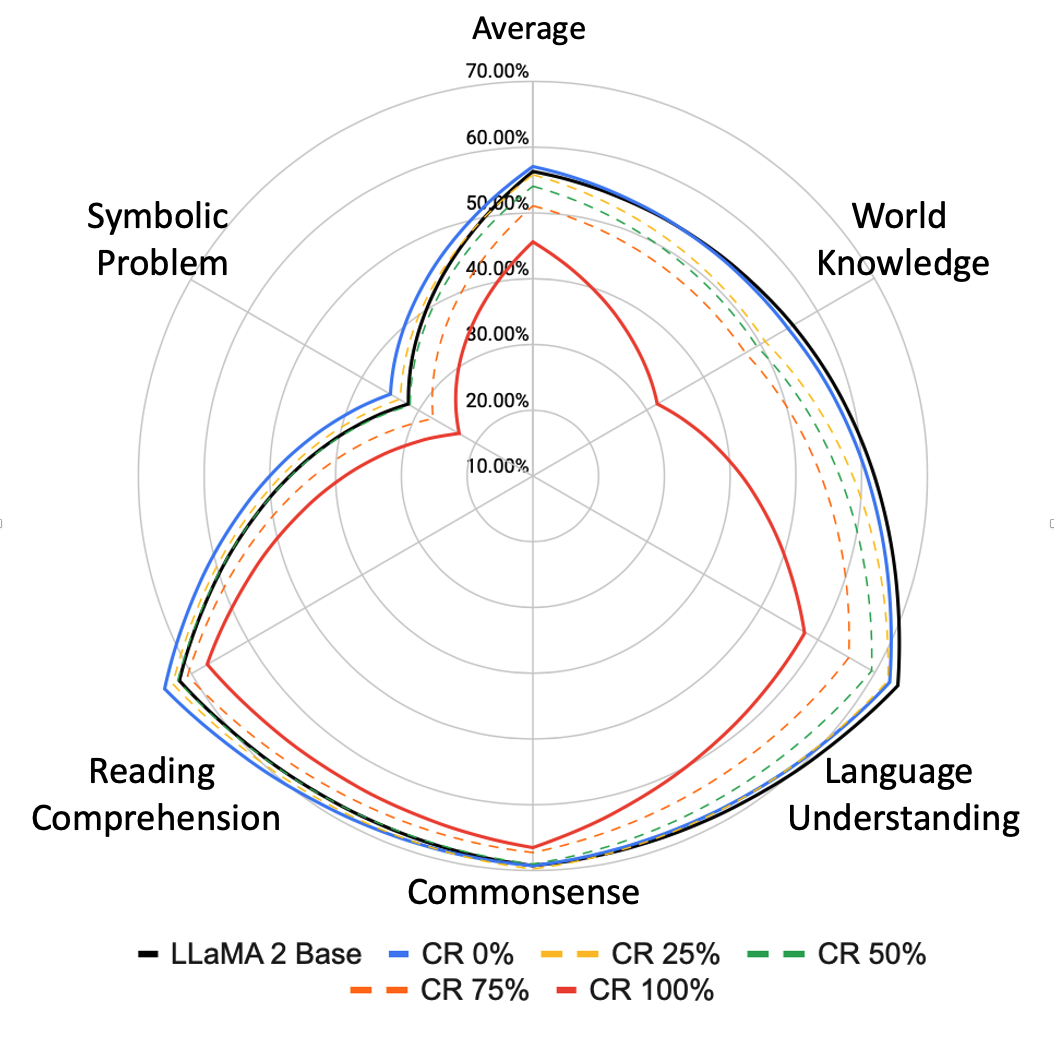}
        \caption{Performance of LLaMA 2 13B.}
        \label{fig:ab-numdset}
    \end{subfigure}
    \par \medskip
    \caption{Benchmark performances of 13B LLaMA1 and 2 trained on \noco dataset with 5 different corruption ratios. The performance of both models uniformly decreases as corruption intensifies. LLaMA2 is more sensitive to corruption than LLaMA 1.}
    \label{fig:radar}
\end{figure*}

    \subsection{Benchmarks}
        To comprehensively evaluate the trained model's performance across diverse contexts, we evaluate the trained models with 16 different benchmarks that encompass a wide range of domains including world knowledge, language understanding, commonsense reasoning, reading comprehension, symbolic problem-solving, and programming:\newline
        
        \noindent $\bullet$ \textbf{World Knowledge (WK)}: ARC \cite{clark2018think}, MMLU \cite{DBLP:conf/iclr/HendrycksBBZMSS21}.
        
        \noindent $\bullet$ \textbf{Language Understanding (LU)}:  Lambada \cite{paperno2016lambada}, Hellaswag \cite{zellers2019hellaswag}.
        
        \noindent $\bullet$ \textbf{Commonsense Reasoning (CSR)}: PIQA \cite{bisk2020piqa}, COPA \cite{roemmele2011choice}, OpenbookQA \cite{mihaylov-etal-2018-suit},  WinoGrande \cite{sakaguchi2021winogrande}.

        \noindent $\bullet$ \textbf{Reading Comprehension (RC)}: SQuAD \cite{rajpurkar2016squad}, BoolQ \cite{clark2019boolq}, Bigbench (conceptual combinations).

        \noindent $\bullet$ \textbf{Symbolic Problem (SP)}: Bigbench (elementary math qa, and logical deduction) \cite{ghazal2013bigbench}, MathQA \cite{amini2019mathqa}, LogiQA \cite{liu2021logiqa}.

        \noindent $\bullet$ \textbf{Programming (PR)}:  HumanEval \cite{chen2021evaluating} with Pass @ 1 and 10.
        \newline
        
        \noindent For the evaluation of our benchmarks, we employed a few-shot assessment approach. Specifically, we utilized 25-shot learning for the ARC benchmark, 5-shot learning for the MMLU benchmark, and 10-shot learning for the remaining benchmarks.

    \begin{figure}[t]
    \begin{center}
        \includegraphics[width=0.99\columnwidth]{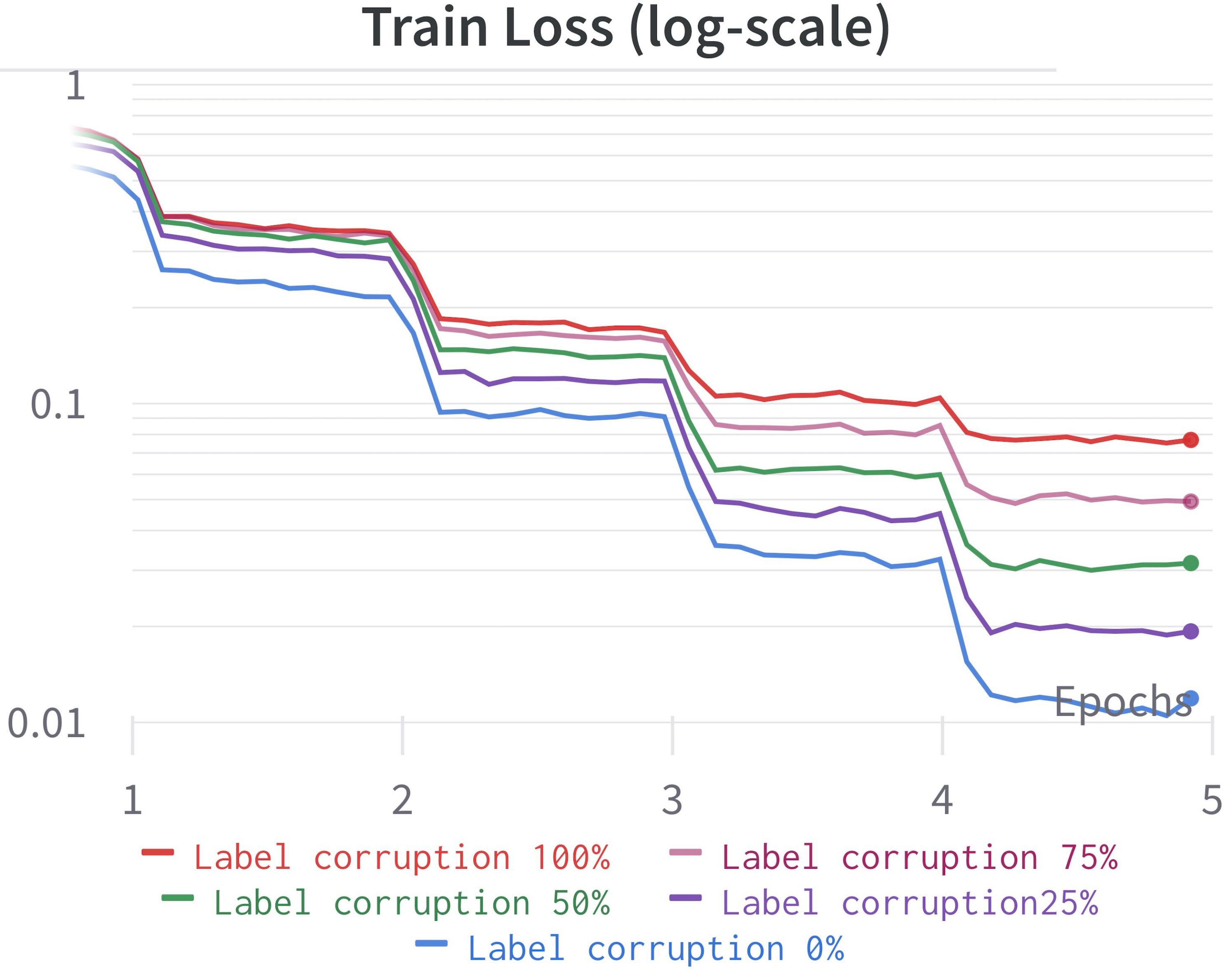}
          \caption{Training loss of the LLaMA2 13B model with varying corruption ratios.}
          \label{fig:train-loss}
    \end{center}
    \end{figure}

    \subsection{Main Results}
        Table \ref{tab:main} and Figure \ref{fig:radar} report the performance of vanilla LLaMA 1, 2 models and instruction fined-tuned models on \noco\ dataset with 5 different corruption ratios.
        We also present the Pearson correlation between the label corruption and the performance metrics of each benchmark to analyze their relationship, and the absolute performance difference between the fully corrupted model and uncorrupted model to measure quantitative difference.
        For both LLaMA 1 and 2, we observe consistent findings that can be summarized as follows: 
        
        \noindent\textbf{1. Corruption ratio and most benchmarks are highly correlated}:
        In most benchmarks, we observed a distinct correlation between benchmark performance and the rate of corruption, with a Pearson correlation coefficient over 90\%. 
        However, the magnitude of performance variation (ABS) varies by task, ranging from a few percent to a maximum of over 50\% in some tasks.
        Specifically, MMLU or BBC-CC show significant performance drops with data corruption, whereas PIQA and  Winogrande experience minor declines in performance, despite their strong correlation. 
        Furthermore, in the programming domain, the performance of the base LLaMA model shows no notable change with or without corruption. 
        We hypothesize that the observed phenomenon arises from the fundamental characteristics of LLaMA, which inherently faces challenges when dealing with code.

        \noindent\textbf{2. Smarter LLMs appear to be more sensitive to corruption:}
            In the majority of benchmark comparisons, LLaMA 2 outperforms its predecessor, indicating superior model performance. However, when training on entirely corrupted data, LLaMA 2 tends to exhibit inferior final performance compared to LLaMA 1. Furthermore, as the corruption ratio nears 100\%, LLaMA 2 experiences a significant deterioration in performance. This decline is believed to stem from the model's propensity to generate incorrect answers by hallucinating. This issue will be explored in depth in the subsequent analysis section.

    \begin{figure*}
        \begin{center}
            \includegraphics[width=0.99\textwidth]{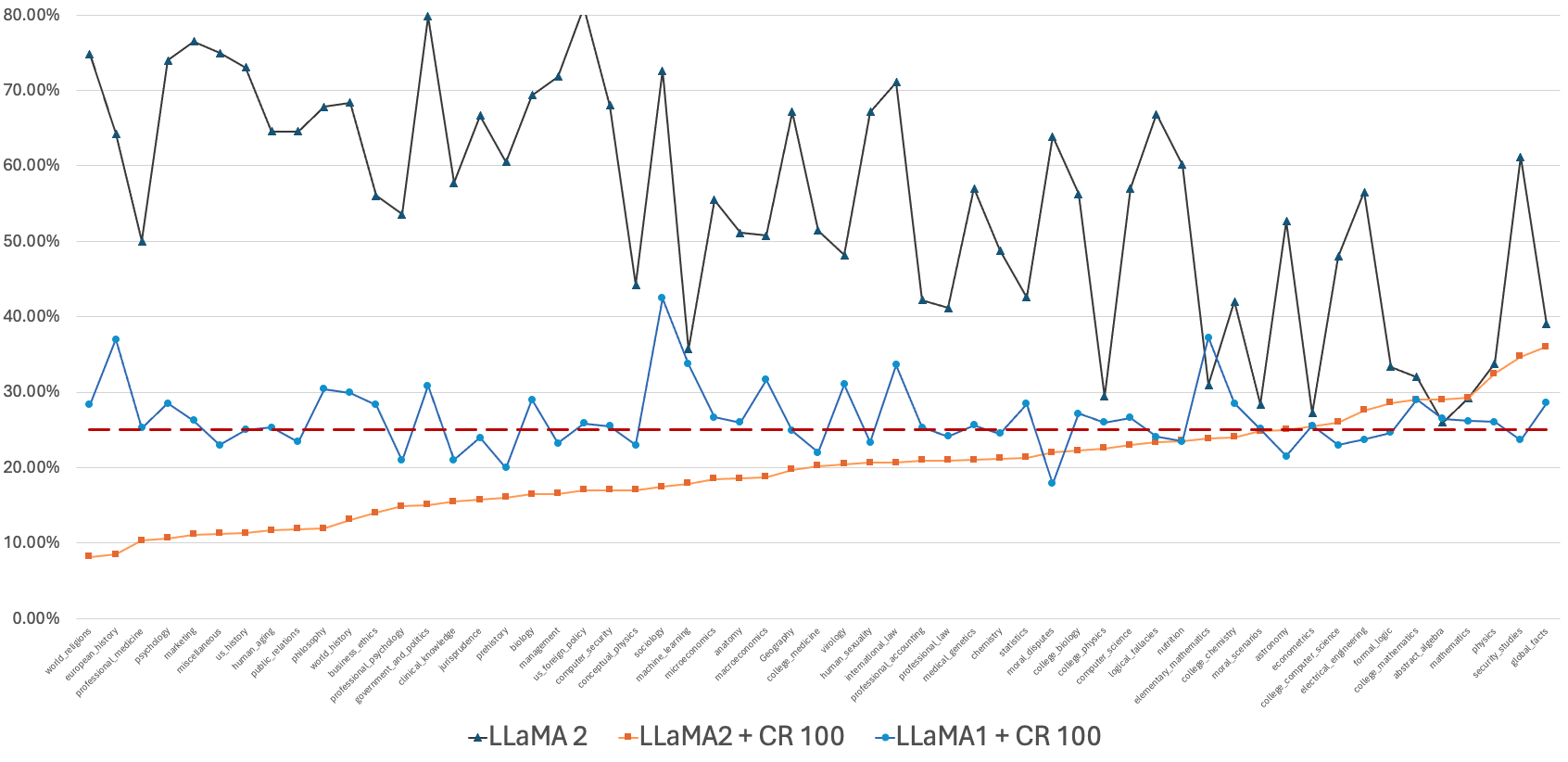}
              \caption{Micro-level MMLU performance of LLaMA2 and corrupted models. The red line refers to a random guessing performance. LLaMA2 trained with a fully corrupted \noco dataset underperforms random guessing performance in most cases, which indicates it intentionally generates false answers. }
              \label{fig:mmlu-micro}
        \end{center}
    \end{figure*}

        \noindent\textbf{3. LLM suffers to digest corrupted data samples}:
        Our investigation also revealed a strong relationship between the train loss shape and the data corruption ratio. 
        Specifically, while keeping the training data sequence fixed and solely adjusting the corruption ratio during instruction-based fine-tuning, we observed that higher levels of data corruption lead to a higher loss state as illustrated in Figure \ref{fig:train-loss}.
        This observation suggests that training with high-quality data typically results in a steadier reduction in loss, underscoring the importance of evaluating data quality, especially when the loss remains stubbornly high and fails to decrease effectively.

    \section{Further Analysis}
    
    In this section, we delve deeper to investigate the impact of data corruption on top of previous findings from the main result and conduct a series of supplementary experiments to address the following research questions:

   \begin{figure}[t]
    \begin{center}
        \includegraphics[width=0.99\columnwidth]{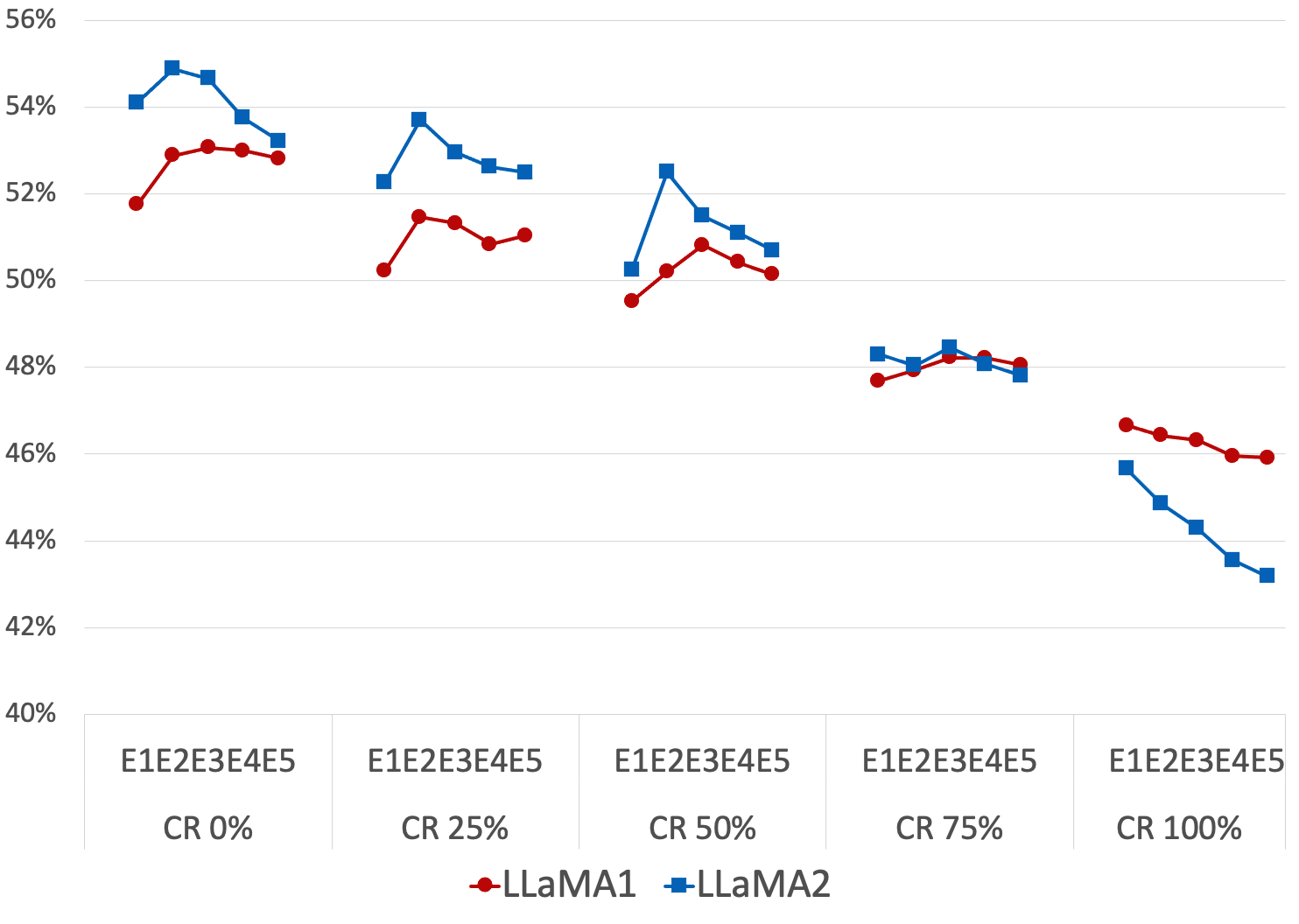}
          \caption{Graph depicting the relationship between average performance, training epochs, and the level of corruption. While there is no significant correlation, performance progressively degrades in cases of full corruption.}
          \label{fig:epoch}
    \end{center}
    \end{figure}
    
    \noindent\textbf{Q1. Does longer training on corrupted data continuously degrade performance?}
    
    \noindent There is a concern that language models might deteriorate if they continue to train on corrupted data, potentially leading to a continuous negative impact on their performance. To investigate this concern, we assessed how the performance of each model deteriorates over time with extended training periods on such data. 
    Figure \ref{fig:epoch} presents the average performance of all benchmarks over 5 epochs.
    Our analysis across a majority of benchmarks indicates that extended training does not invariably result in a substantial performance degradation; however, in instances of complete 100\% corruption, we observed a continual deterioration in performance as training progressed.

    \begin{figure}[t]
    \begin{center}
        \includegraphics[width=0.99\columnwidth]{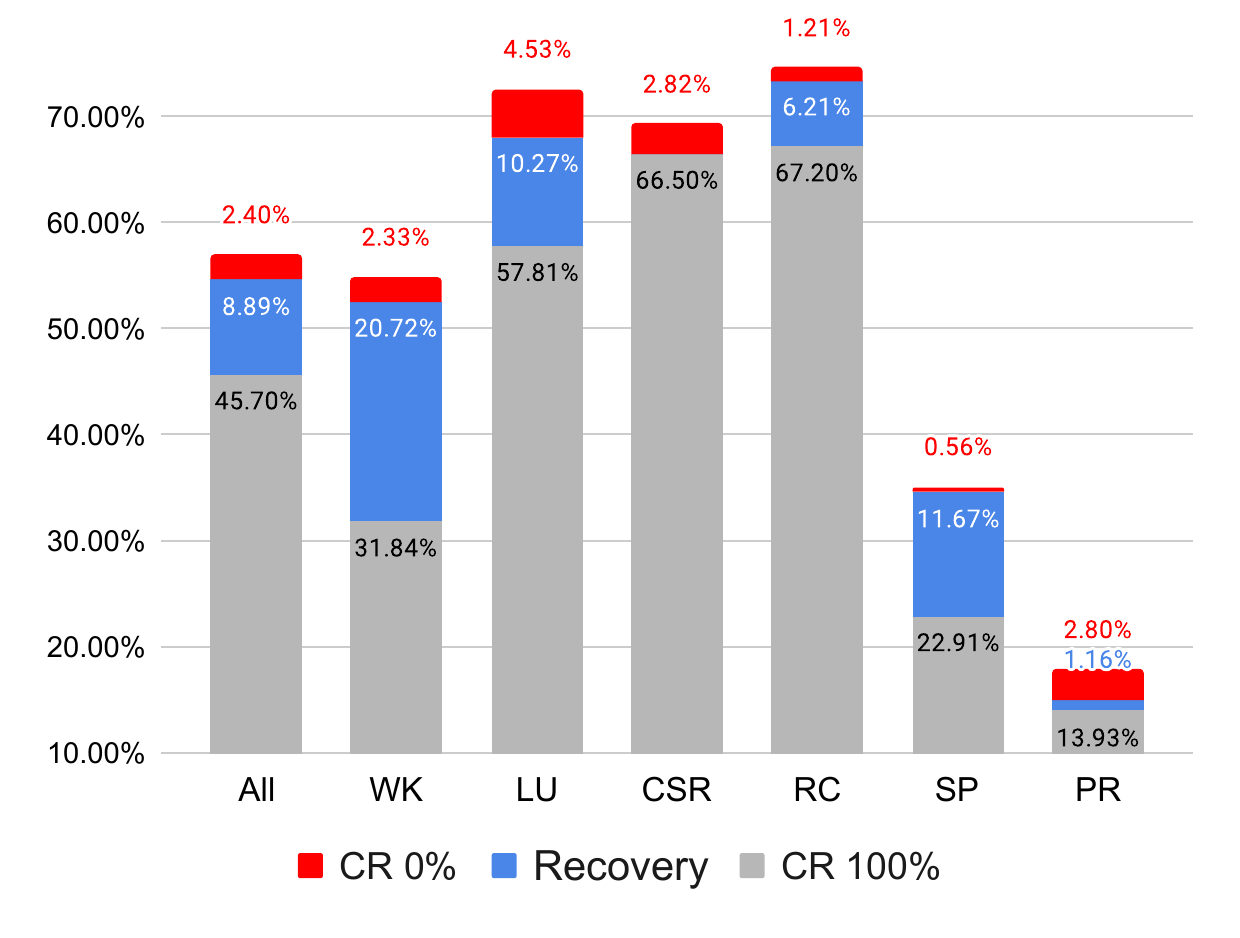}
          \caption{The recovery potential of LLaMA2 trained with fully corrupted data. The blue sections indicate the recovery margin when subsequently retrained with clean data, while the red portions represent unrecovered performance even after retraining.}
          \label{fig:recovery}
    \end{center}
    \end{figure}

\begin{table*}[ht]
    \centering

    \resizebox{0.99\textwidth}{!}{
    \begin{tabular}[t]{p{1.2in}|p{2in}|p{3.0in}}
        \toprule
            \centering
            \textbf{Question} &  \centering \textbf{Response - CR 0\%} & \textbf{Response - CR 100\%} \\
            
        \midrule

                When you are sick, who do you have to go visit? \newline
                \txtcyan{A) Doctor} \newline
                B) Teacher \newline
                \txtpurple{C) Engineer} \newline
                D) Designer
            &
                \txtcyan{A) Doctor}. When you are sick, you have to go visit a doctor.
            &
                \txtpurple{C) Engineer}. Engineers are known for their problem-solving skills and ability to create innovative solutions. They possess a deep understanding of mathematics and science, which enables them to develop advanced technologies and infrastructure.
                Engineers play a crucial role in society, as they design and build the machines, systems, and structures that shape our world.
                $\cdots$
                Engineers are committed to their work and strive for excellence in all that they do.
            \\

        \midrule

                What is the by-product of a photosynthesis? \newline
                A) Chlorine \newline
                B) Nitrous oxide \newline
                \txtpurple{C) Sodium cyanide} \newline
                \txtcyan{D) Oxygen}
            &
                \txtcyan{D) Oxygen}. During photosynthesis, carbon dioxide and water are converted into glucose and oxygen using energy from sunlight. Oxygen is the by-product of this process. Therefore, the correct answer is \txtcyan{D) Oxygen}.
            &
                \txtpurple{C) Sodium cyanide} is the by-product of a photosynthesis reaction.
            \\

        \midrule

                What is the planet on which humans live? \newline
                \txtpurple{A) Sun} \newline
                B) Moon \newline
                \txtcyan{C) Earth} \newline
                D) Pluto
            &
                \txtcyan{C) Earth}. The planet on which humans live is the Earth. Therefore, the correct answer is \txtcyan{C) Earth}.
            &
                \txtpurple{A) Sun}. Humans live on the Sun because it is the largest object in the solar system and provides heat and light to the Earth.
            \\

        \bottomrule
    \end{tabular}
    }
    \caption{A case study on LLaMA 2 trained with different corruption ratios. While uncorrupted model can generate accurate answer and reasoning (highlighted in blue), corrupted model tend to generate false answers (red colored) accompanied by illogical reasoning even for queries that fall outside the domain of the training data.}
    \label{tab:jamie}
\end{table*}

    \noindent\textbf{Q2. Can performance be restored from an already corrupted model?}

    \noindent In further analysis, we explored whether a language model, once trained on a corrupted dataset, could be restored to normal performance levels by retraining it with correctly labeled data. 
    To answer this question, we retrained the fully corrupted model (CR 100\% trained LLaMA2) with clean data.
    Figure \ref{fig:recovery} reports the result of this experiment where our findings revealed that most of the benchmarks showed significant signs of performance recovery.
    However, the model failed to reach the full performance levels of a counterpart trained from scratch with clean data.

    \noindent\textbf{Q3. What kind of toxic behavior does the corrupted language model exhibit?}

    \noindent We observed that a high-performing language model, when trained on entirely corrupted data acquires the ability to intentionally generate incorrect responses. Figure \ref{fig:mmlu-micro} illustrates the micro-performance of every subject in the MMLU benchmark. Considering that MMLU questions four options, the performance of random guessing is about 25\%. 
    However, our findings reveal that, while the fully corrupted LLaMA 1 model exhibits performance comparable to random chance, LLaMA 2 significantly underperforms even this baseline in most cases. Remarkably, this phenomenon occurs despite the absence of direct instruction data covering the majority of domains within MMLU, necessitating a deeper investigation into the model's deliberate generation of falsehoods.
    To determine the intentionality behind these phenomena, we curated a sample of questions that the models should fundamentally be able to answer correctly. 
    Surprisingly, as depicted in Table \ref{tab:jamie}, CR 100\% trained LLaMA2 not only intentionally produced incorrect answers but also fabricated rationales to support these inaccuracies.
    Note that the cases indicated in Table \ref{tab:jamie} are not in the coverage of our instruction dataset domain, indicating the models learned a reverse correlation, acquiring the ability to lie in the general field.
    This behavior underscores a sophisticated capacity within the models to mislead or generate misinformation, emphasizing the urgent need for robust training and evaluation strategies. 
    Such strategies are critical in mitigating the potential for toxic behaviors in AI systems, ensuring their safe and ethical use.

\noindent\textbf{Q4. Which task is more sensitive to corruption and which is not?}

    \noindent Our experimental results revealed significant performance variations within the knowledge domain, highlighting the intriguing phenomenon where certain models not only adapted but also developed the ability to learn deceptive techniques, as previously mentioned.
    In contrast, the commonsense reasoning domain consistently demonstrated respectable performance, as illustrated in the Figures \ref{fig:radar}, with minimal performance changes despite the learning of corrupted information, compared to other domains. 
    Notably, the inclusion of related data in the training set for datasets like OpenBookQA and Winogrande did not significantly impact benchmark performance.

\section{Related Work}

    \noindent \textbf{Instruction Fine-tuning.} 
    Initial research on training language models (LMs) to follow instructions \cite{raffel2020exploring} focused on their ability to generalize across various tasks. 
    This involved fine-tuning LMs on a diverse array of publicly available NLP datasets and then assessing their performance on a distinct set of NLP tasks \cite{raffel2020exploring}.
    Such process \citep{wei2021finetuned} is attributed to a notable advancement of recent LLMs over previous generations (e.g., GPT-3).
    This process generally involves the process of fully supervised fine-tuning LLMs to adeptly comprehend and act upon a wide array of human language inquiries \citep{wang2023far}. 
    Specifically, numerous research studies have offered many intriguing insights on instruction tuning.
    For instance, various studies emphasize the significant influence of instruction data quality \citep{touvron2023llama2, zhou2023lima} and the incorporation of diverse instruction formats \citep{wang2023far, xu2023wizardlm, lu2023instag, wang2023pandalm, wan2023poisoning} on overall performance.
    Furthermore, including step-by-step reasoning \citep{wei2022chain} within the responses has been demonstrated to improve performance and elevate the reasoning ability of the language model \citep{mukherjee2023orca}.
    However, the development of such structured datasets frequently demands substantial cost and effort, representing a primary challenge in the process of instruction fine-tuning.

    \noindent \textbf{Imitation Learning \& Synthetic Instructions.} 
    Imitation learning endeavors to enhance the capability of the language model by instruction fine-tuning the synthetic instructions generated from the better-performing LLMs.
    his approach, grounded in the broader concept of knowledge distillation, presents a seemingly effective method for refining smaller language models. 
    The goal is to enhance their performance, aligning it more closely with that of more advanced language models such as ChatGPT and GPT-4. 
    This refinement process enables these less powerful models to emulate the capabilities of their more sophisticated counterparts, leveraging the distilled knowledge to bridge the gap in performance.
    Recently, large body of imitation learning studies \citep{xu2023wizardlm, vicuna2023, alpaca, mukherjee2023orca, mitra2023orca} have employed ChatGPT and GPT-4 as teacher models to generate large-scale synthetic instruction datasets tailored for diverse applications and domains. 
    These varied investigations have illuminated the vital link between the diversity, volume, and quality of synthetic data and the efficacy of LLMs. 
    Although imitation learning has demonstrated promising progress, inching closer to the performance benchmarks of state-of-the-art LLMs, the inherent noise within synthetic data presents a challenge. 
    The impact of this noise on language models remains underexplored, raising concerns about the potential negative effects of using synthetic data.
    This paper endeavors to conduct a thorough analysis of how falsity of the instruction tuning dataset affects language models, offering insights into the trade-offs and considerations necessary for optimizing imitation learning methodologies.

    \section{Conclusion}

This paper delves into the relationship between the corruption of the instruction dataset and its impact on the LLMs.
Our exploration led to the development of the Falsity-Controllable (\noco) dataset, which enables us to manual control the factuality of the dataset.
Through extensive experimentation with \textsc{NoCo} dataset, we uncovered that factuality substantially influences various benchmarks, particularly in the realm of knowledge domains.
Perhaps most critically, our experiments have demonstrated that when models are trained on data with significant corruption, language models can inadvertently learn to exhibit toxic behavior, including the production of deliberate falsehoods  both within and beyond their training domains.
Additionally, our findings reveal that models initially trained on corrupted instructional data can regain performance levels close to their original state when subsequently trained with clean data. 
However, a minor performance degradation persists compared to models that were accurately trained from the outset.
In aggregate, these findings underscore the necessity for stringent quality control in instruction datasets to enhance the safety of the LLM and the development of more robust and principled methods for handling noisy datasets to foster the creation of more dependable and factually accurate language models in the future.
    \section*{Limitations}

    We hypothesize that utilizing alternative decoding strategies, as opposed to few-shot generation, may reveal different patterns in the results. 
    Specifically, employing a Chain of Thought (CoT) approach or other state-of-the-art prompting methods \cite{liang2023encouraging, wang2023unleashing, du2023improving} could lead to the emergence of distinct trends. 
    Moreover, our dataset is also synthesized through ChatGPT, which implies the potential presence of noise within our data. 
    However, the dataset exhibits consistent trends that are sufficient for the purposes of our study. Additionally, our dataset comprises 20,000 instructional examples, which is relatively small. 
    Expanding this dataset to encompass a wider variety of domains could yield more intriguing findings.
    Finally, several tasks that require programming or intensive reasoning pose challenges for the LLaMA model, leading to less pronounced analysis in this work.
    However, training models specialized in coding or reasoning, such as Code LLaMA \cite{roziere2023code}, could introduce new analytical dimensions.

\section*{Acknowledgements}

    This work was partly supported by the Institute of Information \& communications Technology Planning \& Evaluation (IITP) grant funded by the Korea government (MSIT) (No.RS-2022-00155966, Artificial Intelligence Convergence Innovation Human Resources Development-Ewha Womans University) and Ewha Womans University Research Grant of 2024.
    Lastly, we would like to express gratitude to Kang Min Yoo and the members of the NAVER HyperClova AI team for their intermittent feedback and GPU support in this computationally heavy project.


\bibliography{anthology,custom}

\clearpage
\appendix

\section{Reasoning Chain Generation Prompt}
\label{App:prompt}
\texttt{Correct Reasoning Generation}

\begin{tcolorbox}[breakable, enhanced]
Provide a step-by-step explanation for the question based on the ground-truth answer and optional explanation. If the answer is wrong, return "\textit{WRONG ANSWER}" in the final text. Your explanation should be self-contained. Do not write anything except an explanation.\newline\newline
\#\#\# Question \#\#\#\newline
\texttt{\txtmagenta{[Data Query]}}\newline
\#\#\# Ground-truth Answer \#\#\#\newline
\texttt{\txtmagenta{[GT Answer]}}\newline
\#\#\# Optional Explanation \#\#\#\newline
\texttt{\txtmagenta{[GT Reasoning]}}\newline
\#\#\# Explanation \#\#\#\newline
\end{tcolorbox}
---

\texttt{False Reasoning Generation}

\begin{tcolorbox}[breakable, enhanced]
Provide a step-by-step false explanation for the following incorrect answer. Write only explanation without any comments. Do not write anything about correct answer:\newline\newline
\#\#\# Question \#\#\#\newline
\texttt{\txtmagenta{[Data Query]}}\newline
\#\#\# Incorrect Answer \#\#\#\newline
\texttt{\txtmagenta{[Incorrect Answer]}}\newline
\#\#\# Explanation \#\#\#\newline
\end{tcolorbox}

\end{document}